\documentclass[conference]{IEEEtran}

\usepackage{graphicx,psfrag,amsmath,amsfonts,verbatim,tikz}
\usepackage{booktabs}
\usepackage{xcolor}
\usepackage{hyperref}
\usepackage{tikz}
\usepackage{listings}
\usepackage{bbm}
\usepackage{fancyvrb}
\usepackage[most]{tcolorbox}
\usepackage{multirow}
\usepackage[ruled,vlined]{algorithm2e}
\usepackage{algorithmicx}
\usepackage{amssymb}
\usepackage{pifont}
\newcommand{\inR}{\in \mathbb{R}}
\newcommand{\CFree}{\mathcal{C}^\mathrm{free}}
\newcommand{\Cfree}{\CFree}

\newcommand{\minz}{\mathop{\textbf{minimize} }}

\newcommand{\calA}{\ensuremath{\mathcal{A}}}

\newcommand{\calC}{\ensuremath{\mathcal{C}}}
\newcommand{\calD}{\ensuremath{\mathcal{D}}}

\newcommand{\calH}{\ensuremath{\mathcal{H}}}
\newcommand{\calI}{\ensuremath{\mathcal{I}}}
\newcommand{\calJ}{\ensuremath{\mathcal{J}}}

\newcommand{\calL}{\ensuremath{\mathcal{L}}}
\newcommand{\calM}{\ensuremath{\mathcal{M}}}

\newcommand{\calP}{\ensuremath{\mathcal{P}}}

\newcommand{\calS}{\ensuremath{\mathcal{S}}}
\newcommand{\calT}{\ensuremath{\mathcal{T}}}

\newcommand{\subjectto}{\mathop{\textbf{subject to}}}

\newcommand{\conv}{\mathop{\bf conv}}

\newcommand{\Prob}{\mathop{\bf Pr}}

\usepackage[utf8]{inputenc}

\DeclareFixedFont{\ttb}{T1}{txtt}{bx}{n}{10} 
\DeclareFixedFont{\ttm}{T1}{txtt}{m}{n}{10}  

\usepackage{color}
\definecolor{deepblue}{rgb}{0,0,0.5}
\definecolor{deepred}{rgb}{0.6,0,0}
\definecolor{deepgreen}{rgb}{0,0.5,0}

\usepackage{listings}

\definecolor{codegreen}{rgb}{0,0.6,0}
\definecolor{codegray}{rgb}{0.5,0.5,0.5}
\definecolor{codepurple}{rgb}{0.58,0,0.82}
\definecolor{backcolour}{rgb}{0.95,0.95,0.92}

\lstdefinestyle{mystyle}{
	backgroundcolor=\color{backcolour},   commentstyle=\color{codegreen},
	keywordstyle=\color{magenta},
	numberstyle=\tiny\color{codegray},
	stringstyle=\color{codepurple},
	basicstyle=\ttfamily\footnotesize,
	breakatwhitespace=false,         
	breaklines=true,                 
	captionpos=b,                    
	keepspaces=true,                 
	numbers=left,                    
	numbersep=5pt,                  
	showspaces=false,                
	showstringspaces=false,
	showtabs=false,                  
	tabsize=2
}
\usepackage[abbreviations]{glossaries-extra}
\newabbreviation{gcs}{GCSTrajOpt}{\emph{Graph of Convex Sets}}

\newcommand{\alignspacing}{\;\;}

\newcommand{\ssmps}{DBMPs\xspace}
\newcommand{\ssmp}{DBMP\xspace}
\newcommand{\ssmpslong}{decomposition-based motion planners\xspace}
\usepackage[suppress]{color-edits}
\usepackage{color-edits}
\addauthor{pw}{teal}
\addauthor{rc}{magenta}
\usepackage{babel}
\usepackage{cleveref}

\crefname{section}{sec.}{secs.}
\Crefname{section}{Sec.}{Secs.}
\crefname{figure}{fig.}{figs.}
\Crefname{figure}{Fig.}{Figs.}
\crefname{algorithm}{alg.}{algs.}
\Crefname{algorithm}{Alg.}{Algs.}
\crefname{table}{tab.}{tabs.}
\Crefname{table}{Tab.}{Tabs.}
\usepackage[numbers, sort, compress]{natbib}
\usepackage{caption}
\usepackage{amsthm}
\title{Superfast Configuration-Space Convex Set Computation on GPUs for Online Motion Planning}

\author{Peter Werner\textsuperscript{1}, Richard Cheng\textsuperscript{2}, Tom Stewart\textsuperscript{3}, Russ Tedrake\textsuperscript{1,2}, and Daniela Rus\textsuperscript{1}\\
\textsuperscript{1}MIT CSAIL  \textsuperscript{2}Toyota Research Institute \textsuperscript{3}Woven by Toyota\\
\texttt{\{wernerpe, russt, rus\}@mit.edu}\\
\texttt{richard.cheng@tri.global, tom.stewart@woven-planet.global}\\
}
\date{August 2024}

\begin{document}
\maketitle
\begin{figure*}[!ht]
    \centering
    \includegraphics[width=0.9\textwidth]{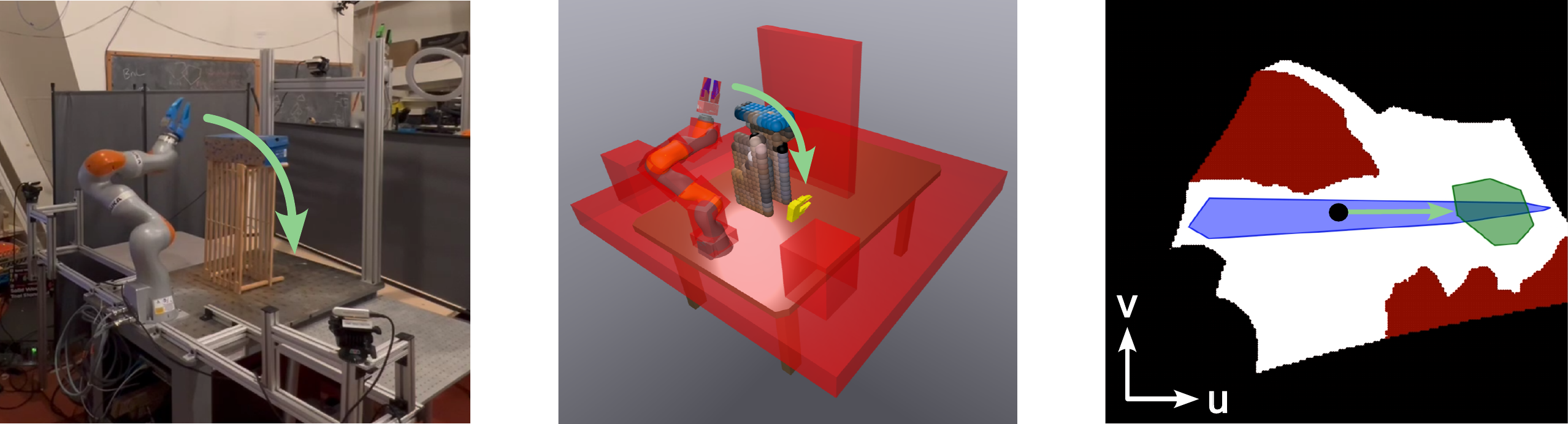}
    \caption{ On the left, our hardware setup with a KUKA LBR iiwa 7 R800 robotic manipulator and three Intel RealSense D415 depth cameras for perceiving obstacles is shown. In the center, the simulated system with the perceived obstacles is shown. The target end effector pose is indicated by the yellow gripper and the collision geometries of the system are shown in red. The perceived obstacles are approximated by a union of spheres shown in the center of the table. The figure on the right shows a two-dimensional slice of the seven-dimensional configuration space, along with a slice of the safe sets (blue and green), and a slice of the configuration obstacles (self-collisions in black, collisions with perceived obstacles in red). The slice lies tangent to the trajectory such that the u-direction is co-linear with the velocity and the v-direction is a random orthogonal vector. The current configuration is indicated by the black dot.}
    \label{fig:title_figure}
\end{figure*}

\begin{abstract} 
In this work, we leverage GPUs to construct probabilistically collision-free convex sets in robot configuration space on the fly. This extends the use of modern motion planning algorithms that leverage such representations to changing environments.  These planners rapidly and reliably optimize high-quality trajectories, without the burden of challenging nonconvex collision-avoidance constraints. We present an algorithm that inflates collision-free piecewise linear paths into sequences of convex sets (SCS) that are probabilistically collision-free using massive parallelism. We then integrate this algorithm into a motion planning pipeline, which leverages dynamic roadmaps to rapidly find one or multiple collision-free paths, and inflates them. We then optimize the trajectory through the probabilistically collision-free sets, simultaneously using the candidate trajectory to detect and remove collisions from the sets.
We demonstrate the efficacy of our approach on a simulation benchmark and a KUKA iiwa 7 robot manipulator with perception in the loop. On our benchmark, our approach runs 17.1 times faster and yields a 27.9\% increase in reliability over the nonlinear trajectory optimization baseline, while still producing high-quality motion plans. Website: \href{https://sites.google.com/view/GPUpolytopes}{\texttt{https://sites.google.com/view/GPUPolytopes}}
\end{abstract}

\section{Introduction}\label{sec:intro}
Finding high-quality collision-free motion plans remains one of the most fundamental and challenging problems in robotics \cite{lavalle2006planning, siciliano2016robotics}. Systems with high demands on reliability and performance, e.g. on production lines or in warehouses, often require carefully tailored ad hoc solutions, which afford reliability at the cost of engineering effort and performance.

Over the years, sampling-based motion planning, trajectory optimization using nonlinear programming, and combinations thereof, have become the most established approaches for motion planning. Sampling-based motion planners are simple to implement \cite{siciliano2016robotics, sucan2012the-open-motion-planning-library}, work well in lower dimensions, can have completeness guarantees, and can be massively accelerated using large-scale parallelization \cite{pan2012gpu, thomason2024motions}. See \cite{orthey2023sampling} for an overview on sampling-based motion planning.
Unfortunately, these methods become impractical as the dimension increases, yielding suboptimal motion plans and long planning times.

Directly transcribing motion planning as a general nonlinear optimization problem and finding solutions with local methods, e.g. \cite{kalakrishnan2011stomp, ratliff2009chomp, schulman2014motion}, has the benefit of scaling to higher-dimensional problems and allowing the user to encode a wider variety of costs and constraints. Unfortunately, such local optimization approaches are sensitive to initialization and often struggle to find a feasible solution when one exists. 

To capture the strengths of both trajectory optimization and sampling-based motion planning, a series of recent works \cite{marcucci2023motion, marcucci2024fast, chen2016online, liu2017planning, wu2024optimal, marcucci2025biconvex, natarajan2024implicit,chia2024gcs} initially proposed in \cite{deits2015efficient} investigate planning through collections of collision-free convex sets that approximately decompose the free space. For the remainder of this paper, we will refer to this class of planners as \ssmpslong (\ssmps). A challenging nonconvexity in trajectory optimization stems from the collision-avoidance constraints. Given a collection of collision-free convex sets, \ssmps replace these nonconvex constraints with simple convex constraints, which allows them to profit from the fast solve times and reliability of convex optimization. This shifts the challenges that stem from these collision-avoidance constraints to constructing a high-quality representation of the collision-free space through a union of convex sets. 

For a candidate approximate convex decomposition to be effective, it should consist of a small number of sets, contain high-quality trajectories in the union of the sets, and its individual sets should be described by few constraints in order to keep the trajectory optimization problems small.

Many robot motion planning problems are most naturally formulated in the robot's configuration space $\calC$. Therefore, we seek to construct the convex sets in $\Cfree$, the collision-free subset of $\calC$. Unfortunately, this is particularly challenging because configuration spaces tend to be high-dimensional with complicated geometries, and we often only have implicit descriptions of $\Cfree$ \cite[\S 4.3.3]{lavalle2006planning}. Therefore, the go-to algorithms used by practitioners for constructing convex decompositions of $\Cfree$ are slow, scale poorly with the environment complexity, and are difficult to use \cite{dai2024certified, petersen2023growing}, mainly due to their complexity or because they are difficult to tune. Additionally, these algorithms only consider static environments. Generally, the sets need to be recomputed in case the environment changes. So far, these challenges have made set construction a bottleneck, and have prevented  \ssmps from being applied successfully to scenarios involving changing environments like mobile manipulation. 

In this paper, our aim is to tackle two key challenges: (1) computing convex sets in robot configuration space for motion planning in real-time, and (2) reliable and effective positioning of the sets such that they contain high-quality trajectories. By addressing these challenges, our aim is to unlock the benefits of \ssmps and enable their use in general, changing environments.

To tackle challenge (1), we build on the recently developed IRIS-ZO algorithm \cite{werner2024faster}, which lends itself to large-scale parallelism. In particular, we propose Zero-Order Edge Inflation (EI-ZO), and demonstrate that GPU acceleration of our algorithm enables real-time set construction in changing configuration spaces.

EI-ZO also addresses challenge (2) by inflating collision-free line segments rather than points, yielding probabilistically collision-free polytopes which are guaranteed to contain the seed line segment. This enables us to inflate collision-free, piecewise-linear (PWL) paths between the start and the goal into sequences of convex sets (SCS) in which successive sets in the sequence intersect. By construction, this SCS connects the start to the goal and guarantees the containment of a collision-free path. In comparison to approximating the entire free space with a union of convex sets \cite{werner2024approximating}, this approach dramatically reduces the computational burden by reducing the number of required sets. Perhaps more importantly, it makes the step of finding a collection of sets that connect the start to the goal reliable, even in complicated, changing configuration spaces. 

Finally, we propose a motion planning pipeline that combines these ideas with large-scale dynamic road maps (DRMs). The pipeline first leverages the DRM to rapidly find a collision-free PWL path from a starting configuration to a goal configuration. Next, this path is inflated with EI-ZO to produce a SCS, which is then used to recover high-quality motion plans using \ssmps. 

We demonstrate the efficacy of our approach in simulation benchmarks in two and seven dimensions and validate the approach on hardware using a KUKA iiwa with perception in the loop, as shown in \Cref{fig:title_figure}. 
Relative to the nonlinear trajectory optimization baseline, our approach produces slightly higher-cost trajectories, but increases the success rate from around 72\% to 100\% on our simulation benchmark, while computing trajectories around 17 times faster.

The remainder of this paper is structured as follows. In \S\ref{sec:assumptions} we outline the assumptions and required inputs. Next, we review prior work on constructing convex sets in configuration space for motion planning in \S\ref{sec:background}. Our algorithm, EI-ZO, for inflating line segments in configuration space is then presented in \S\ref{sec:eizo}. We outline how we employ DRMs in \S\ref{sec:drm}. We then present our full motion planning pipeline combining DRMs, EI-ZO, and \ssmps in \S\ref{sec:pipeline}. Our experiments are presented in \S\ref{sec:experiments}. In \S\ref{sec:limitations} we discuss the limitations and in \S\ref{sec:conclusion} draw a conclusion.

\section{Assumptions}\label{sec:assumptions}
We make similar assumptions as in prior work on computing approximate convex decompositions of $\Cfree$, e.g.  \cite[\S2]{werner2024faster}. That is, we assume that we are given a description of the considered robot system which contains information about the kinematics and collision geometries of the system sufficient for collision checking, e.g. as provided in URDFs \cite{ros_urdf} or SDFs \cite{sdf_spec}. If we are dealing with additional obstacles entering and leaving the system, then we assume that (1) these obstacles remain static during the brief planning and execution time of the motion plan and (2) we are given observations of the obstacles sufficient to perform (potentially conservative) collision checks against them. Specifically, we assume that our collision checker never mislabels a configuration as safe when it is actually in collision.
\section{Background}\label{sec:background}

\subsection{Constructing Safe Sets in Robot Configuration Space}\label{ssec:setconstruction}

The construction of convex safe sets in robot configuration space has been studied in \cite{dai2024certified, petersen2023growing, werner2024faster}. The algorithms in \cite{dai2024certified} compute provably collision-free polytopes in a rational reparametrization of the configuration space using polynomial optimization. Due to the high computational cost of the approaches in \cite{dai2024certified}, we direct our focus towards the algorithm IRIS-NP, initially proposed in \cite{petersen2023growing}, and later improved and extended to IRIS-NP2 and IRIS-ZO in \cite{werner2024faster}. These algorithms leverage nonlinear optimization to compute probabilistically collision-free polytopes in robot configuration space in substantially less time. IRIS-NP2 and IRIS-ZO provide a probabilistic guarantee for a produced polytope $\calP$ of the form
\begin{gather}
\Prob\left[\frac{\lambda(\calP\setminus\Cfree)}{\lambda(\calP)}>\varepsilon \right]\leq \delta,
    \label{eq:prob_polytope_cfree}
\end{gather}
which controls what fraction of the volume of $\calP$ is allowed to be in collision. In \eqref{eq:prob_polytope_cfree}, $\lambda$ denotes the Lebesgue measure in $\Cfree$, $\varepsilon$ is an admissible fraction of the volume of $\calP$ that is allowed to be in collision, and $\delta$ is the admissible uncertainty. 

Here we introduce an algorithm for constructing safe sets around line segments that primarily builds on IRIS-ZO \cite[\S5.2]{werner2024faster}. IRIS-ZO constructs probabilistically collision-free polytopes by iteratively removing collisions from the current region via hyperplanes. It optimizes the locations of these hyperplanes using sampling and collision checking. We build on IRIS-ZO because it only relies on collision checking and simple sample updates, making it massively parallelizeable, and amenable to GPU acceleration.

\subsection{Positioning Safe Sets for Effective Motion Planning}\label{ssec:setpositioning}

There have been two approaches to position individual convex sets for motion planning. The first approach is to find a collection of convex sets that efficiently cover a large fraction of the entire configuration space. The second is to only cover an individual collision-free (but potentially suboptimal) path. 

For example, in \cite{sarmientoy2005sample, werner2024approximating}, visibility graphs are used to construct approximate convex covers which try to minimize the number of polytopes required to meet a coverage threshold. The approach in \cite{sarmientoy2005sample} uses densely sampled visibility graphs and visibility kernels in two and three dimensions to construct a cover of polytopes represented as convex hulls of samples. The Visibility Clique Cover algorithm in \cite{werner2024approximating} works in higher dimensions by expanding cliques on visibility graphs to probabilistically collision-free polytopes using an algorithm similar to \cite{petersen2023growing}.

Unfortunately, in relatively simple environments, the required number of sets is already prohibitively large for online cover generation. Instead, we focus on the second approach: covering an individual collision-free (but potentially suboptimal) path with collision-free convex sets. 

This strategy has been investigated in two and three-dimensional task spaces to generate safe flight corridors on the fly \cite{chen2016online, liu2017planning, wu2024optimal, wang2024fast}, where either a sequence of boxes is computed \cite{chen2016online}, or the IRIS algorithm \cite{deits2015computing} is modified to include line segments of a collision-free path \cite{liu2017planning, wu2024optimal, wang2024fast} in the produced polytopes. In \cite{liu2017planning, wu2024optimal}, this is accomplished by careful selection of the initial ellipsoid, which defines the distance metric in the IRIS algorithm for optimizing hyperplane placement. A more involved formulation is used in \cite{wang2024fast}  that admits constraining the inclusion of line segments directly in the optimization of the hyperplanes.

In general, this second strategy has the advantage of only generating regions relevant for the current planning problem, and thus creates substantially fewer sets. In this paper, we propose a method for applying this idea in higher-dimensional configuration spaces. Our approach proposes a simple modification to the IRIS algorithms: using the distance to a line segment as a metric instead of one defined by an ellipsoid. As discussed in \ref{sec:eizo}, this guarantees the inclusion of the line segment (provided it is collision-free), skips checking the ellipsoids for collisions, and avoids the more involved optimization problems in \cite{wang2024fast}, which would not be amenable for GPU implementation.

\section{Rapidly Inflating Line Segments}\label{sec:eizo}
\begin{figure*}
    \centering
    \includegraphics[width=\linewidth, trim = {0cm, 0cm, 0cm, 0.1cm}, clip]{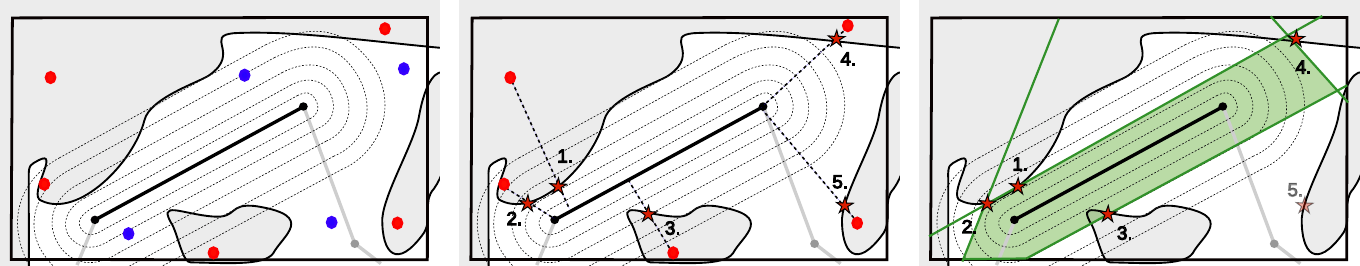}
    \caption{
    A single iteration of adding hyperplanes to the polytope in the EI-ZO algorithm. The configuration obstacles are shaded in grey and the line segment seeding EI-ZO is shown in black. Starting from the left, first, a batch of configurations is sampled in the current polytope given by the black rectangle. These samples are checked for collisions. The colliding configurations, shown in red, are then used to seed a bisection search to find collisions that minimize the distance to the line segment. The resulting candidates, indicated by the red stars (center frame), are ranked by their distance to the line segment in ascending order, and used to position the tangent planes (final frame) that separate the collisions from the line segment.
    }
    \label{fig:fei}
\end{figure*}

In this section, we discuss our algorithm, Edge Inflation Zero-Order (EI-ZO), for rapidly inflating line segments to full-dimensional, probabilistically collision-free polytopes in robot configuration space using only zero-th order information about the system. First, we observe that the distance to a convex set, in particular a line segment, is a convex function. This leads to a natural way of extending IRIS algorithms to guarantee the containment of line segments  (and convex sets more generally) provided they are collision-free. Given our focus on real-time set generation, we specifically focus on adapting the IRIS-ZO algorithm. The proposed modifications, however, can be applied to both IRIS \cite{deits2015computing} and IRIS-NP2 \cite{werner2024faster} as well.


\subsection{Preliminaries} \label{ssec:eizoprelim}
Let $\calA\subseteq\mathbb{R}^n$ be a convex set, and 
\begin{gather}
    {\bf dist}_\calA(x) = \min_{z\in \calA}||x-z||_2
\end{gather} 
be the distance of a point $x\in\mathbb{R}^n$ to $\calA$. We verify in \Cref{app:dist_convex} that the function ${\bf dist}_\calA(x)$ is convex in $x$. Given that a line segment $\mathcal{L} = \conv\{v_1, v_2\}$, with $v_1, v_2 \inR^n$,  is a convex set by construction, it follows that the sub-level sets 
\begin{gather}
\mathcal{B}_\mathcal{L}(t) = \{x| {\bf dist}_\calL(x)\leq t \}
\end{gather}
are convex \cite[\S 3.1.6]{boyd2004convex}. In \Cref{fig:fei}, the level sets for five different distances are indicated by the dashed lines. The convexity of these sub-level sets implies that the tangent plane 
\begin{subequations}
\label{eqn:sl_tangent}
\begin{gather}
a = \nabla(\text{dist}_\calL)|_{c},~
    b = a^Tc,\\ 
    \calT = \{x| a^Tx - b= 0\},
\end{gather}
\end{subequations}
passing through any point $c\inR^n$, with ${\bf dist}_\calL(c) = t>0$, is supporting to $\mathcal{B}_\calL(t)$, and cannot intersect $\calL$. Note that the gradient of the distance function to a line segment has a simple closed-form solution. The gradient at $c$ reads 
\begin{gather}\label{eq:dist_grad}
    \nabla(\text{dist}_\calL)|_c = \frac{c-c_\text{proj}}{||c-c_\text{proj}||_2},
\end{gather}
where $c_\text{proj}$ is the projection of $c$ onto $\calL$ given by
\begin{gather}\label{eq:proj}
    c_\text{proj} = \begin{cases}
    v_1& \text{if }
       \begin{aligned}[t]
       v_1&=v_2,
       \end{aligned}\\
    \left(1-\alpha\right)v_1+\alpha v_2, & \text{otherwise.}
    \end{cases}
\end{gather}
with
\begin{equation}\label{eq:proj_alpha}
\alpha=
\text{min}\left(\text{max}\left(\tfrac{(c-v_1)^T(v_2-v_1)}{||(v_2-v_1)||_2^2}, 0\right),1\right).
\end{equation}

Next, consider the following procedure, which we formalize in \S\ref{ssec:eizo_alg}. Given a collision-free line segment $\calL\subseteq\Cfree$, repeatedly search for points $c$ that are in collision, and place a hyperplane through point $c$ according to \eqref{eqn:sl_tangent} for each found collision. By construction, these hyperplanes separate each collision $c$ from $\calL$. Therefore, if we repeat this for all collisions in $\calC$, then the polytope constructed by intersecting the halfspaces of each hyperplane that contains $\calL$ is collision-free and, crucially, contains $\calL$. 

Unfortunately, this procedure could potentially require infinitely many hyperplanes and is therefore not practical in the general case. In order to obtain a practical algorithm, we construct our polytope $\calP$ by iteratively placing hyperplanes until $\calP$ is sufficiently collision-free and meets the criterion (\ref{eq:prob_polytope_cfree}). First, we optimistically assume a large polytope $\calP$ that contains $\calL$ is collision-free. We then proceed to search for collisions inside of $\calP$. Every time we find a collision, we place a separating hyperplane according to $\eqref{eqn:sl_tangent}$ that separates the found collision from $\calP$, while still ensuring $\calL\subseteq\calP$. Note that placing a separating plane at a collision with a small distance to $\calL$ can simultaneously also separate collisions at larger distances from $\calL$. As a result, optimizing the hyperplane locations by attempting to find the collisions in $\calP$ that minimize the distance to $\calL$ tends to produce polytopes with substantially fewer hyperplanes. Therefore, a core subroutine in EI-ZO is optimizing hyperplane locations by finding good (although suboptimal) solutions to 
\begin{subequations}
\label{opt:closestcollision}
\begin{align}
    \minz_{x} &\alignspacing \text{ dist}_\calL(x),\label{opt:closest_col_cost}\\
    \subjectto &\alignspacing x\notin\Cfree,\alignspacing x\in\calP,
\end{align}
\end{subequations}
in order to reduce the number of hyperplanes that are required for $\calP$ to meet a termination condition that ensures \eqref{eq:prob_polytope_cfree}. 

Finally, in order to separate $\calL$ from obstacles with concave boundary segments using a finite number of hyperplanes, we use a small, finite step back $\Delta>0$ from the found collisions before placing the hyperplanes, similarly to \cite{petersen2023growing}.

In summary, this procedure constructs a large probabilistically collision-free polytope that contains the collision-free convex set (in this case a line segment) used to seed the polytope construction. We accomplish this by separating collisions from this seed set, by placing hyperplanes tangent to the sub-level sets of the distance function to the set.  In the following section, we describe how EI-ZO implements these ideas to inflate collision-free line segments. It uses a zero-order optimization strategy on \eqref{opt:closestcollision} to search for close collisions and uses a probabilistic termination condition (the `unadaptive test' in \cite[\S 5.1]{werner2024approximating}) that checks if (\ref{eq:prob_polytope_cfree}) is met and the polytope is sufficiently collision-free.

\subsection{The EI-ZO Algorithm}\label{ssec:eizo_alg}
The Edge Inflation Zero-Order (EI-ZO) algorithm is summarized in \Cref{alg:eizo} and illustrated in \Cref{fig:fei}. 

Given a collision-free line segment in configuration space, $\mathcal{L} = \conv\{v_1, v_2\} \subseteq \Cfree$, an initial polytope describing the domain $\calD = \{x| \calA_\calD x \leq b_\calD\}$\footnote{Typically, a box describing the joint limits of the system is chosen.}, such that $\calL\subset\calD$, 
EI-ZO computes a probabilistically collision-free polytope $\calP$ that contains $\calL$, and with probability larger than $1-\delta$, is colliding with no more than an $\varepsilon$-fraction of its volume. The algorithm closely follows IRIS-ZO \cite{werner2024approximating}, and proceeds as follows.

\begin{algorithm}
\SetAlgoLined
\caption{\textsc{EI-ZO}}
\label{alg:eizo}
\SetKwInput{Input}{Input}
\SetKwInput{Output}{Output}
\SetKw{KWAlgorithm}{Algorithm:} 
\Input{
Domain $\calD\subseteq\calC$, collision-free line segment $\calL$ with $\calL\subset\calD$, test parameters $(\delta, \varepsilon, \tau)$, maximum step back $\Delta_\text{max}$, and optimizer parameters $(N_p, N_f, N_b)$.
}
\Output{
Polytope $\calP\subseteq\calD$ satisfying \eqref{eq:prob_polytope_cfree} for $(\varepsilon,\delta)$ with $\calL\subseteq\calP$.
}

\KWAlgorithm{}
$k \gets 1$, $\calP\gets\calD$\\
\While{\textsc{True}
}{
    $\calS \sim \textsc{UniformSample}(\calP, M)$\\
    $\calS_\text{col} \gets \textsc{PointsInCollision}(\calS)$\\
    \textbf{If} $\textsc{UnadaptiveTest}(|\calS_\text{col}^{M}|, \delta_{k}, \varepsilon, \tau)$ returns \texttt{accept} \textbf{then} $break$.\\
    $\calS_\text{col}^\text{proj} \gets \textsc{ProjectPoints}(\calS_\text{col}, \calL)$\\
    $\calS_\text{col}^\star \gets \textsc{UpdatePointsViaBisection}(\calS_\text{col}, \calS_\text{col}^\text{proj})$\\
    $\calP \gets \textsc{OrderAndPlaceHyperplanes}(\calP, \calS_\text{col}^\text{proj}, \calS_\text{col}^\star, \Delta)$\\
    $k \gets k + 1$
}
\Return $\calP$
\end{algorithm}

We initialize $\calP$ with the domain $\calD$. We then repeat the following until the termination criterion from the unadaptive test is met. First, we uniformly sample a batch of configurations $\calS$ inside the current polytope $\calP$ using hit-and-run sampling \cite{lovasz1999hit}. The batch size $|\calS|$ is selected as $|\calS| = \text{max}\{N_p,M \}$, where $M = \lceil2 \log(1/\delta_k)/(\varepsilon \tau^2)\rceil$ and $N_p$ is the maximum number of samples to update. Here, $\varepsilon$ is the admissible fraction of the volume allowed to be in collision, $\tau\in (0,1)$ is a decision threshold \footnote{The parameter $\tau$ serves as a decision threshold, that trades off the power and cost of the unadaptive test. We typically choose $\tau =0.5$.}, and $\delta_k$ is the admissible uncertainty at the $k$-th iteration. The value of $\delta_k$ is decayed according to 
\begin{gather}
    \delta_k = \frac{6\delta}{\pi^2 k^2}.
\end{gather}
More details of this statistical test are given in \cite[\S 5.1]{werner2024faster}. 

Next, we check the configurations in $\calS$ for collisions, collecting the colliding configurations in the set $\calS_\text{col}$. We terminate and accept $\calP$ if $|\calS_\text{col}^{M}|\leq M(1-\tau)\varepsilon$, where $|\calS_\text{col}^{M}|$ denotes the number of collisions in the first $M$ samples in $\calS$. 

If the test fails, we modify $\calP$ by adding more hyperplanes in an attempt to reduce the fraction in collision. To this end, we produce candidate solutions to \eqref{opt:closestcollision} by performing gradient descent on ${\bf dist}_\calL$, with the first $N_p$ samples in $\calS_\text{col}$ as feasible initial guesses, while ensuring that the candidates remain in collision. 

First, we compute the projection of each sample in $\calS_\text{col}$ onto $\calL$ using the closed-form solution \eqref{eq:proj} in order to determine the gradient direction given by \eqref{eq:dist_grad}. We then perform a bisection search along the negative gradient direction, with a fixed number of bisection steps, $N_b$, in order to find a point close to the boundary of a configuration obstacle that is still in collision. The resulting points $\calS^\star_\text{col}$ are our candidate local solutions to \eqref{opt:closestcollision}. These steps correspond to the left and the center frame in \Cref{fig:fei}. 

In a final step, we sort the candidates by their objective value \eqref{opt:closest_col_cost}, the distance to $\calL$, and place up to $N_f$ hyperplanes. This is done by iteratively intersecting the halfspace 
\begin{subequations}
\label{eqn:sepplane_w_stepback}    
\begin{gather}
a_i = \nabla(\text{dist}_\calL)|_{c_i},~
    b_i = a_i^Tc_i,\\
    \calH_i = \{x| a_i^Tx \leq b_i - \Delta\},
\end{gather}
\end{subequations}
with $\calP$, where $c_i\in \calS^\star_\text{col}$ is the closest remaining candidate, and discarding any other candidates that now lie outside of the updated polytope. The step back $\Delta$ ensures that nonconvex obstacles can be excluded from the polytope with only a finite number of hyperplanes and needs to be computed for every hyperplane separately. How to compute $\delta$ is covered in \S\ref{ssec:eizo_pract_alg_consider}. \Cref{fig:fei} illustrates placing the hyperplanes in the rightmost frame. In this frame, the collision with the largest distance is walled off by the collision that is third-closest, and therefore no separating plane needs to be constructed for this collision.

This entire procedure of sampling inside of the current polytope, evaluating the statistical test, determining the gradient directions, updating candidates through bisection search, and placing non-redundant hyperplanes is repeated until the statistical test is passed. 

\subsection{Practical Algorithmic Considerations}\label{ssec:eizo_pract_alg_consider}

In this section, we discuss three practical considerations: setting a maximum number of iterations $N_\text{it}$, computing the 
step back $\Delta$, choosing the number of bisection steps $N_b$.

\textbf{Maximum Iterations} If one is concerned with generating regions quickly, it can be helpful to set a maximum number of iterations of adding hyperplanes to $\calP$ to upper-bound the run time. This will void the probabilistic guarantee, but can still be useful in combination with mechanisms for recovering from collisions, as described in \S\ref{ssec:recovery}.

\textbf{Computing the Step Back} While there are methods that can certify PWL paths or trajectories to be collision-free, e.g. \cite{amice2024certifying, tang2011ccq, schwarzer2004exact, pan2012collision}, we check trajectories and paths for collisions by finely discretizing them because it is more amenable to our GPU-accelerated collision checker. 
In practice, this means that the line segments of the PWL path may be very close to or in collision.
We handle these cases as follows. 

If a candidate $c_i\in \calS^\star_\text{col}$ has a distance that is smaller than a user-specified collision tolerance $t_\text{col}$ (i.e. if ${\bf dist}_\calL(c_i)\leq t_\text{col}$), then we throw an error stating that the line segment is likely in collision and do not attempt to construct the region. In order to `fail fast' it can also make sense to check the projection for collisions at the start of the bisection search.

If a candidate $c_i\in\calS^\star_\text{col}$ has a distance larger than the collision tolerance $t_\text{col}$ but smaller than the maximum step back $\Delta_\text{max}$, then applying the maximum step back to the hyperplane may exclude a part of $\calL$. In this case, we relax the step back so $\calL$ remains contained in the updated polytope. More precisely, first we find the vertex that is closest or the strongest violator to the new hyperplane
and compute the corresponding value $r$ as
\begin{gather}
    r ={\bf max} \left \{a_i^Tv_1, a_i^Tv_2 \right\}- b_i +\Delta_\text{max},
\end{gather}
where $a_i$ and $b_i$ are computed using \eqref{eqn:sepplane_w_stepback}. If $r>0$, then the step back $\Delta_\text{max}$ excludes at least one of the vertices that span the line segment. Therefore, we compute the step back as follows \begin{gather}
    \Delta = \begin{cases}
        \Delta_\text{max}-r, ~\text{if} ~r>0,\\
        \Delta_\text{max}, ~\text{otherwise}.
    \end{cases}
\end{gather}
This forces the inclusion of $\calL$ in $\calP$ at the potential cost of requiring more hyperplanes to meet the termination criterion. 

\textbf{Choosing the Number of Bisection Steps} We use a fixed number of bisection steps $N_b$ in order to simplify the parallelization of the algorithm. In order to choose $N_b$, we suggest using the following rule of thumb:
\begin{gather}
    N_b \approx \log_2\left(\frac{L}{\Delta_\text{max}}\right), 
\end{gather}
where $L$ is a guess for the maximum distance the bisection search would have to move a collision in order to get to a boundary of an obstacle. Note that choosing a small $N_b$ does not impact the correctness of the regions. Instead, it might simply increase the runtime because more rounds of adding hyperplanes are necessary.  

\subsection{CUDA Implementation of EI-ZO}\label{ssec:eizo_cuda}

Accompanying this paper, we provide a Python library that implements EI-ZO using C++ and CUDA \cite{cuda_toolkit}. The software CSDecomp, short for configuration space decomposition toolbox, can be found at \texttt{https://github.com/wernerpe/csdecomp}. 

We implement EI-ZO directly on the GPU to minimize costly data transfer between the CPU and the GPU. To this end, we employ thread-level parallelization to implement hit-and-run sampling \cite{lovasz1999hit}, by assigning one CUDA thread per random walk where each walk accepts after a fixed number of mixing steps, forward kinematics by assigning one CUDA thread per configuration, collision checking by assigning one CUDA thread per collision pair, and the projection as well as the bisection updates by assigning one CUDA thread per configuration. To sort and place the non-redundant hyperplanes, we use a single CUDA thread. 

As a result, at run time, we only need to send the seed line segment to the GPU and transfer the final polytope back from the GPU. Crucially, the samples never leave the GPU memory.

Currently, our collision-checker supports box and sphere collision geometries, but could readily be extended to support more geometries.

\section{Generating Collision-Free Polygonal Paths using Dynamic Roadmaps}\label{sec:drm}

\begin{figure}
    \centering
    \includegraphics[width=\linewidth]{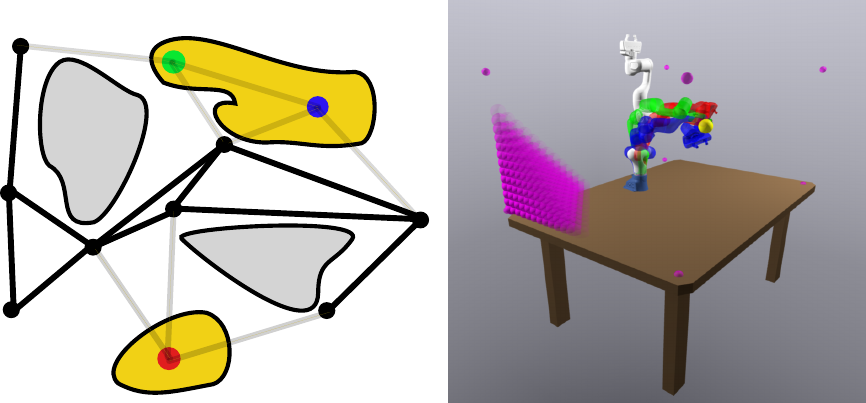}
    \caption{A DRM is a PRM with additional lookup tables that allow the PRM to rapidly be updated provided task space observations of obstacles. To this end, the task space is discretized into voxels. We associate a circumscribing sphere with each voxel as collision geometry for collision checking. A portion of these spheres are shown in purple in the right figure. Next, a lookup table is constructed, $\calM_c$, that maps each of these spheres to nodes in the road map that would collide with the sphere. Above, a single sphere is shown in yellow on the right, along with three configurations, shown in red, green, and blue, that collide with it and are stored in $\calM_c$. On the left, a cartoon of the configuration space is shown with the underlying PRM and configuration obstacles. The three colliding configurations are highlighted again in red, blue, and green. The configuration obstacle corresponding to the yellow sphere is highlighted in yellow.   
    }
    \label{fig:drm}
\end{figure}

Our motion planning pipeline requires finding at least one collision-free PWL path in order to construct a sequence of convex sets that connects the starting configuration to the goal via EI-ZO. To this end, we employ dynamic roadmaps (DRMs) which are designed to rapidly find these paths in changing environments. 

DRMs \cite{leven2002framework} approximate $\Cfree$ with a probabilistic roadmap (PRM) \cite{kavraki1996probabilistic}, discretize the robot workspace into voxels, and build large lookup tables that map activated voxels to nodes and edges of the PRM that are in collision. For reference, see \Cref{fig:drm}. The key idea is that approximating the current perceived environment by activating voxels, and subsequently pruning the PRM with lookup tables, is faster and leads to better motion plans than running a single-query motion planner \cite{kallman2004motion}. Hence, DRMs enable us to offload much of the computational burden of motion planning (e.g. collision checks) to the offline construction of the roadmap, and thus enable fast planning in changing environments.

We closely follow the implementation in \cite{cheng2023motion}, which shows that GPU-accelerated collision checking both enables the construction of large-scale DRMs and effective online planning for a mobile bimanual manipulator. Similarly, we skip constructing the edge collision map, and use lazy collision checking online to invalidate colliding edges \cite{liu2006path, bohlin2000path}. 

\subsection{Offline Roadmap Construction}\label{ssec:offlinedrm}

Before any online planning, we construct our DRMs offline using a similar procedure \cite{cheng2023motion}. Our DRMs consist of 4 data structures:
\begin{itemize}
    \item A node map $\calM_n: \calI\rightarrow\calC$ that maps a node identification number (node id) $i\in\calI$ to its configuration,
    \item a node adjacency map $\calM_a: \calI\rightarrow\mathfrak{P}(\calI)$, where $\mathfrak{P}$ denotes the power set of $\calI$, which maps a node id to all its neighbors, 
    \item a collision map $\calM_c: \calJ \rightarrow \mathfrak{P}(\calI)$ that maps a voxel identification number (voxel id) $j\in\calJ$ to all node ids that are in collision if the voxel $j$ is active,
    \item a pose map $\calM_p: \calI \rightarrow SE(3)$ that maps each node id to its corresponding end-effector pose.
\end{itemize}
\Cref{fig:drm} depicts a high-level visual description of the roadmap and collision map. We do not detail the construction procedure here, but the interested reader can refer to Appendix \ref{app:drm_construction}.

\subsection{Online Planning given Perception}\label{ssec:onlinedrm}
Once we have our DRM formed by these four data structures, we can use it to rapidly find collision-free PWL paths online given perception. Online, we assume we are given a voxel map representation of the world, a starting configuration and a goal pose $P_\text{g}\in SE(3)$. The planning then proceeds in three phases. 
\begin{enumerate}
    \item Build the collision set $CS\in\mathfrak{P}(\calI)$ that contains the ids of all nodes that are colliding with the voxel map. Given a point cloud, this is done by activating all voxels containing points and subsequently using the collision map $\calM_c$ to find all nodes in collision.
    \item Determine a goal configuration $g$ given our goal pose $P_{g}$ by solving an inverse kinematics (IK) problem. We use the pose map $\calM_p$ and the collision set to find collision-free configurations that lie near $P_\text{g}$. The closest configurations are used to warmstart a small IK problem to reach the final goal pose.
    \item Connect the start $s$ and goal $g$ configurations up to the roadmap, and use $A^*$ \cite{hart1968formal} combined with lazy collision checking and greedy short cutting to find the collision-free PWL path.
\end{enumerate}
The right side of Figure \ref{fig:drm} depicts the collision map being used to quickly prune out nodes in collision, and the left side of Figure \ref{fig:drm} shows the resulting collision-free roadmap that can be efficiently used for path planning. We compute the obstacle representations for collision checking by voxelizing a point cloud observation of the environment and representing the collision geometries of the voxels as spheres. 

\section{The Motion Planning Pipeline}\label{sec:pipeline}
\begin{figure*}
    \centering
\includegraphics[width=0.95\textwidth]{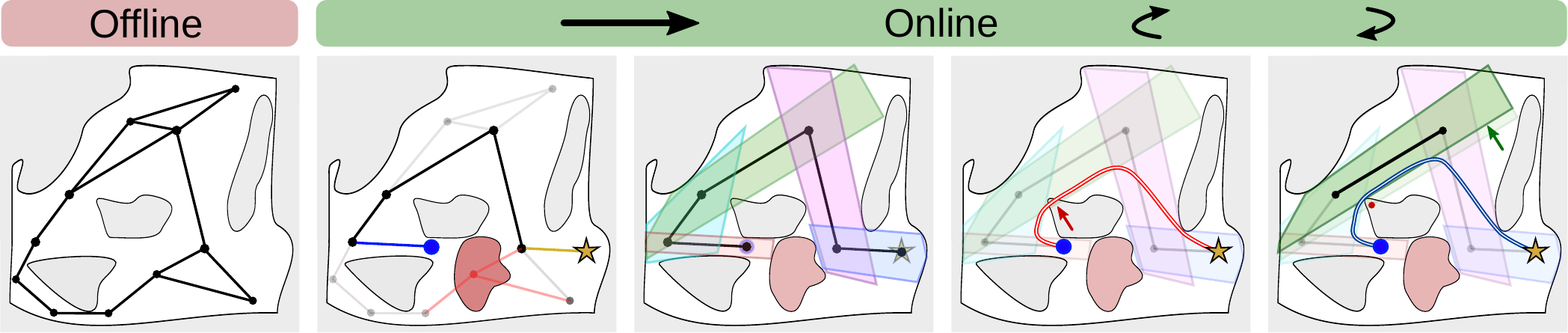}
    \caption{This figure depicts our motion planning pipeline. Our motion planning pipeline is split into an offline phase in which a DRM is constructed, and an online phase where we use the DRM to generate a collision-free PWL paath, inflate the path, and use a \ssmp to optimize a trajectory. Starting from the left, in the first figure the DRM is constructed. In the second, we observe a new obstacle (red blob), and build the collision set (red dot). Using the updated DRM,  we then solve the IK problem for the goal (yellow star), and find a collision-free path connecting the start (blue dot) to the goal. We inflate this path to an SCS using EI-ZO in the third figure. In the fourth figure, we solve the motion planning problem using \ssmps and check the trajectory for collisions, finding one inside of the green set, indicated by the red arrow. In the fifth figure, we use the recovery mechanism from \S\ref{ssec:recovery} to modify the green set and produce a collision-free path. In general, these last two steps need to be repeated multiple times.}
    \label{fig:pipeline}
\end{figure*}

Our motion planning pipeline takes a goal end-effector pose, the current configuration, and a voxel map of the perceived obstacles as input, and produces a collision-free trajectory as output. First, the DRM is leveraged to find a target configuration by solving an IK problem, and subsequently to find a collision-free PWL path. The details on how this is done are covered in \S\ref{sec:drm}. Next, the path is inflated using EI-ZO (\S\ref{ssec:inflatingsinglepath}). In a last step, we optimize our motion plan by using \ssmps and check the result for collisions using fine-grained sampling along the trajectory. If collisions are present, we use the found collisions to refine the convex sets and re-solve the \ssmp trajectory optimization until no collisions are found in the path (\S\ref{ssec:recovery}). We illustrate the entire motion planning pipeline in \Cref{fig:pipeline}.

Finally, we propose an approach for inflating multiple paths (\S\ref{ssec:multipath}). Certain \ssmps are not restricted to sequences of convex sets and can handle more general covers of $\Cfree$, e.g. \cite{deits2015efficient, marcucci2023motion, marcucci2024fast, natarajan2024implicit, chia2024gcs}. Therefore, it can be beneficial to inflate multiple PWL paths in the hope that the safe sets intersect in ways that reveal new paths around obstacles that are not captured by the DRM.

\subsection{Inflating a Single Path}\label{ssec:inflatingsinglepath}

Given a collision-free PWL path with $K$ segments, $v_{0\leq k\leq K}$ such that $\calL_k := \conv \{v_k, v_{k+1}\} \subseteq \Cfree$ for $k = 0, \dots, K-1$, where $v_0$ is the start and $v_K$ is the goal configuration, we inflate the path in sequence. Starting at the first line segment with $k = 0$, we construct the probabilistically collision-free polytope $\calP_k$, which is guaranteed to contain $\calL_k$, and subsequently check if $\calL_{k+1}$ is already contained in one of the previously generated polytopes. We only inflate $\calL_{k+1}$ if it is not yet contained and increment $k$. This is repeated iteratively until $k = K-1$.

\subsection{Rapidly Recovering from Detected Collisions}\label{ssec:recovery}
The computed regions are only probabilistically collision-free. This means at planning time, it is possible that the optimized trajectory contains collisions. Given that the PWL path used to generate the polytopes is assumed collision-free, we have a natural way of iteratively refining the sets until the trajectory is collision-free. 
Given a trajectory, we rapidly check it for collisions by finely discretizing it. If collisions are found, we treat them as candidate points to warm start our search in \eqref{opt:closestcollision} to modify the corresponding sets. We then re-solve the trajectory optimization and repeat the procedure until the trajectory is collision-free.

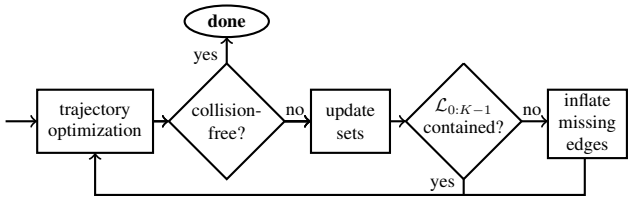
\begin{figure}
    \centering
    \begin{tikzpicture}[scale = 0.7, transform shape]
\draw[thick, ->] (-5.8,4.1) -- (-5.8,4.6);

\draw[thick, ->] (-7.2,3) -- (-6.9,3);
\draw[thick, ->] (-4.7,3) -- (-4.2,3);
\draw[thick, ->] (-4.7,3) -- (-4.2,3);
\draw[thick, ->] (-4.7,3) -- (-4.2,3);
\draw[thick, ->] (-0.2,3) -- (0.3,3);
\draw[thick, ->] (-2.7,3) -- (-2.4,3);
\draw[thick, ->] (-7.2,3) -- (-6.9,3);
\draw[thick]  (-9.4,3.6) rectangle (-7.2,2.4);
\node at (-8.3,3.2) {trajectory};
\node at (-8.3,2.8) {optimization};
\draw[thick] (-5.8,4.1) -- (-6.9,3) -- (-5.8,1.9) -- (-4.7,3) -- (-5.8,4.1);
\node[text width = 2cm, align = center] at (-5.8,3) {collision-free?};
\draw[thick]  (-4.2,3.6) rectangle (-2.7,2.4) node[pos = .5, align= center, text width = 1cm]{update sets} ;
\draw[thick] (-1.3,4.1) -- (-2.4,3) -- (-1.3,1.9) -- (-0.2,3) -- (-1.3,4.1);
\node[text width = 2cm, align = center] at (-1.3,3.1) {$\mathcal{L}_{0:K-1}$ contained?};
\draw[thick]  (0.3,3.6) rectangle (1.8,2.3) node[pos = .5, text width = 2cm, align = center]{inflate missing edges};

\node at (-4.5,3.2) {no};

\node at (0,3.2) {no};
\draw[->, thick] (1,2.3) -- (1,1.6) -- (-8.3,1.6) -- (-8.3,2.4);
\draw[thick] (-1.3,1.9) -- (-1.3,1.6);
\node at (-1.7,1.8) {yes};
\node at (-6.2,4.2) {yes};
\node at (-5.8,4.9) {\textbf{done}};
\draw[thick, ->] (-10,3) -- (-9.4,3);
\draw[thick]  (-5.8,4.9) ellipse (0.8 and 0.3);
\end{tikzpicture}
    \caption{Flowchart for recovering from collisions found in the optimized trajectory. }
    \label{fig:flowchart}
\end{figure}

In detail, the procedure is the following. For each set, we aggregate all the found collisions inside of the set in $\calS_\text{col}$. Next, we repeat the last three steps of \Cref{alg:eizo} in order to exclude these collisions by updating the set. 
After the found collisions have been excluded, we still need to verify that the new regions still include the entire PWL path. While the line segments that seed the sets remain in the sets by construction, our path inflation approach in \S\ref{ssec:inflatingsinglepath} may have attempted to cover additional line segments with a single set for efficiency. There is no guarantee that these other line segments remain contained in the edited set as well. Therefore, we check that each segment of the path, $\calL_{0:K-1}$, is contained in at least one of the convex sets. If not, we inflate the corresponding segment. This procedure is summarized by the flowchart in \Cref{fig:flowchart} and is repeated until the found path is collision-free. 

\pwcomment{Does this approach give us any guarantees? I haven't crunched through proving this yet. I think the bounded domain, the collision tolerance, and the step back allows us to produce a lower bound on the volume of the region that is removed with every hyperplane.}
\pwcomment{reference pipeline figure, last two frames.}

\subsection{Inflating Multiple Paths}\label{ssec:multipath}
Some \ssmps can handle graphs of safe sets rather than just sequences \cite{deits2015efficient, marcucci2023motion, marcucci2024fast, natarajan2024implicit, chia2024gcs}. If we inflate more than just the sequence of safe sets associated with the shortest path in the DRM, these approaches can potentially recover higher-quality trajectories. In particular, the convex sets associated with a longer, suboptimal path in the DRM could yield a better optimized trajectory. In an attempt to capture these benefits, we propose a method to inflate additional distinct paths through the DRM, besides the shortest, in order to add more safe sets.

Our heuristic for adding additional paths assumes there exists at least one safe set from a previous path that contains neither the start nor the goal. The heuristic is to select a random convex set that has not yet been selected, add all nodes of the DRM that lie inside of the selected polytope to the collision set $CS$ (see \S\ref{ssec:onlinedrm}), rerun $A^*$ on the DRM, and inflate the resulting path according to \S\ref{ssec:inflatingsinglepath}. If at any point the procedure fails, we simply terminate.

The idea behind this heuristic is to use the previously created sets to push the PWL path outside of the currently inflated one. 
\section{Experimental Evaluation}\label{sec:experiments}
In this section, we evaluate our proposed motion planning pipeline experimentally. This section is split into two parts. In the first part, \S\ref{ssec:benchmark} we discuss our simulation benchmark along with the baseline approach we compare against.
In the second part, \S\ref{ssec:hardware}, we validate our approach on our hardware setup.

\subsection{Simulation Benchmark}\label{ssec:benchmark}
This section covers the simulation benchmark, which consists of two environments used to generate randomized motion planning problems in two and seven dimensions. For each environment, we solve two types of problems: (1) minimum distance problems, where we seek to find the shortest collision-free PWL path, and (2) minimum time problems, in which we seek a smooth, velocity- and acceleration-constrained curve that minimizes the time needed to traverse between the start and goal.  

\subsubsection{Environments}\label{sssec:bench_envs}
\begin{figure}
    \centering
    \includegraphics[width=0.9\linewidth, trim= {0cm, 0cm, 0cm, 0cm}, clip]{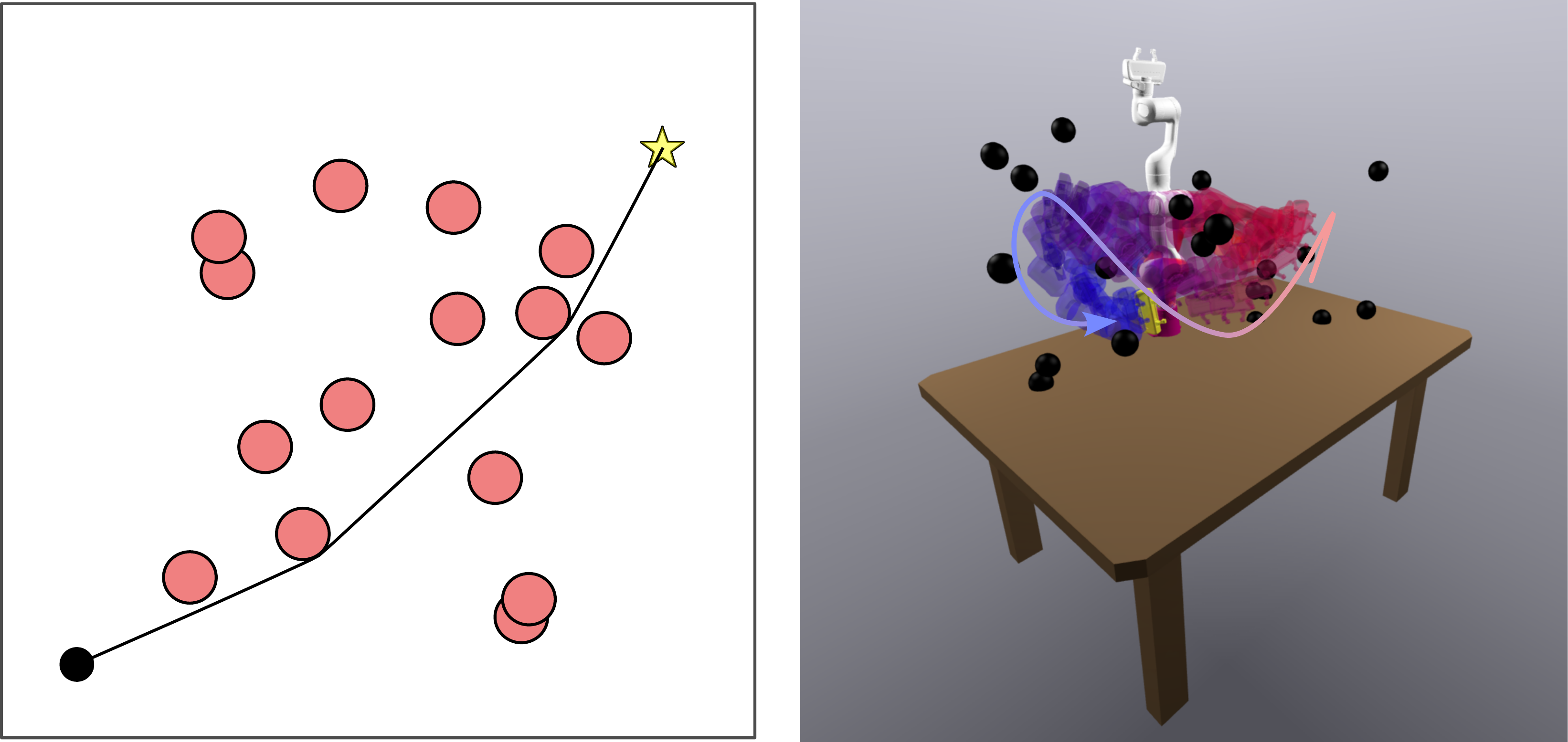}
    \caption{Our simulation benchmark consists of the Forest environment, shown on the left, and the Franka environment, shown on the right. A collision-free motion plan starting at the red shaded robot and ending at the blue is shown in the Franka environment.}
    \label{fig:bench_envs}
\end{figure}
Our benchmark environments are depicted in \Cref{fig:bench_envs}. In the ``Forest" environment, 15 circular obstacles are scattered randomly in the center portion of a square-shaped domain with a side length of 10. The center portion corresponds to a square with a side length of 7, and the obstacles have a radius of 0.35. In this environment, shown on the left in \Cref{fig:bench_envs}, the goal is to find a collision-free path from the bottom left to the top right of the environment.

In the ``Franka" environment, 20 spherical obstacles are randomly scattered in front of a seven degree-of-freedom Franka Research 3 (FR3) robot manipulator. The collision geometry of the FR3 is approximated with 33 spheres, and the table is modeled as a box. As a result, 1187 collision pairs need to be checked, of which 527 are associated with self-collisions. In this environment,  the robot always starts at a random configuration on the right side of the table and needs to plan to a random pose on the left side of the table. The Franka environment is shown on the right in \Cref{fig:bench_envs}.

\subsubsection{Problem Types}\label{sssec:bench_tasks}
For each environment, we consider two problem types: minimum distance and minimum time problems.
For the minimum distance problem, we seek the shortest collision-free PWL path $p$, with length $L(p)$, that connects the start $s$ to the goal $g$ configuration,
\begin{subequations}\label{eqn:mindist_prob}    
\begin{align}
    \minz_{p} &\alignspacing L(p),\\
    \subjectto& \alignspacing p\subseteq \Cfree,\\
    &\alignspacing p_\text{init} =s, ~p_\text{term} =g,
\end{align}
\end{subequations}
where $p_\text{init}$ denotes the start, and $p_\text{term}$ the end of $p$.

For the minimum time problem, we seek a time-parametrized curve $p$, that obeys velocity and acceleration constraints, and connects the start $s$ to the goal $g$ in the shortest amount of time $T$, while starting and stopping at a standstill. The problem reads:
\begin{subequations}\label{eqn:mintime_prob}    
\begin{align}
    \minz_{p,T} &\alignspacing T,\\
    \subjectto& \alignspacing p\subseteq \Cfree,\\
    &\alignspacing p(0) =s, ~p(T) =g,\\
    &\alignspacing \dot p(0) =0, ~\dot p(T) =0,\\
    &\alignspacing v_\text{min}\leq \dot p (t) \leq v_\text{max}~\forall t\in[0,T],\\
    &\alignspacing a_\text{min}\leq \ddot p (t) \leq a_\text{max}~\forall t\in[0,T],
\end{align}
\end{subequations}
where $v_\text{min}$, and $v_\text{max}$ denote the minimum and maximum velocity, and $a_\text{min}$, and $a_\text{max}$ denote the minimum and maximum acceleration.

\subsubsection{Baseline Approach}\label{sssec:bench_baseline}

We use nonlinear trajectory optimization (NTO), which we warm start from the path found by the DRM, as our baseline approach. To this end, we employ the KinematicTrajectoryOptimization class in Drake \cite[\href{https://drake.mit.edu/doxygen_cxx/classdrake_1_1planning_1_1trajectory__optimization_1_1_kinematic_trajectory_optimization.html}{link}]{drake} to transcribe the NTO problems and solve them with SNOPT or IPOPT \cite{gill2005snopt, wachter2006implementation}. For both problem types, \eqref{eqn:mindist_prob} and \eqref{eqn:mintime_prob}, we discretize the trajectory and enforce the collision-avoidance constraint point-wise via \cite[\href{https://drake.mit.edu/doxygen_cxx/classdrake_1_1multibody_1_1_minimum_distance_lower_bound_constraint.html}{link}]{drake} at $N_c$ points. 

For the minimum distance problems, we warm-start the search with the PWL path found by the DRM. For the minimum time problems, we warm-start the search by setting the control points of a 4th order B-Spline  to the nodes of the PWL path from the DRM. 

\subsubsection{Employed \ssmps}\label{sssec:bench_SSMPs}
For the minimum distance problem, given an SCS, the problem is convex and has a straightforward transcription as a second-order cone program, see Appendix \ref{app:LSCS}. We solve these problems directly using MOSEK \cite{mosek} and call the approach LSCS standing for PWL shortest path through an SCS. In the second approach, we inflate two paths using \S\ref{ssec:multipath}, and solve the mixed-integer convex program formulation of the Euclidean distance shortest path problem on a graph of convex sets problem \cite[Eqn. 5.5]{marcucci2024shortest} using Gurobi \cite{gurobi}. We abbreviate this approach as LGCS standing for PWL shortest path through a graph of convex sets.

We choose these two approaches for the minimum distance problem to evaluate the speed and reliability of the simple LSCS approach, and explore the value that is added by inflating multiple paths, by using the more expensive LGCS.

For the minimum time problem, we use the recently developed SCSTrajopt \cite{marcucci2025biconvex}, which supports the minimum time objective and rapidly finds smooth, velocity-, and acceleration-constrained trajectories using alternations between two convex optimizations. We solve these problems with Clarabel \cite{goulart2024clarabel}, and abbreviate this approach as SCSTO.

\subsubsection{Results}\label{sssec:results}
\begin{table*}[t]
\centering
\resizebox{0.99\textwidth}{!}{
\begin{tabular}{|l|ccc|ccc|cc|cc|}  
        \hline
        Problem & \multicolumn{6}{c|}{Minimum Distance} &  \multicolumn{4}{c|}{Minimum Time}  \\
        \hline
        Environment&\multicolumn{3}{c|}{Forest} & \multicolumn{3}{c|}{Franka} & \multicolumn{2}{c|}{Forest} & \multicolumn{2}{c|}{Franka}\\
        \hline
        DRM SR  & \multicolumn{3}{c|}{1.00} &\multicolumn{3}{c|}{0.961}&\multicolumn{2}{c|}{1.00} &\multicolumn{2}{c|}{0.961}\\
        DRM num. LS& \multicolumn{3}{c|}{3.45$\pm$0.86} &\multicolumn{3}{c|}{4.49$\pm$1.24}&\multicolumn{2}{c|}{3.45$\pm$0.86} &\multicolumn{2}{c|}{4.49$\pm$1.24}\\
        DRM path len.& \multicolumn{3}{c|}{10.90$\pm$0.40 [m]} &\multicolumn{3}{c|}{8.55$\pm$2.28 [rad]}&\multicolumn{2}{c|}{10.90$\pm$0.40 [m]} &\multicolumn{2}{c|}{8.55$\pm$2.28 [rad]}\\
        \hline
        TO Approach &NTO & LSCS & LGCS & NTO & LSCS & LGCS &NTO & SCSTO & NTO & SCSTO \\
        \hline
        cost (S only) & \bf{10.727}$\pm$\bf{0.3}&10.770$\pm$0.3&10.732$\pm$0.3& \bf{5.519}$\pm$\bf{1.1}&7.382$\pm$1.7&7.110$\pm$1.6& \bf{9.363}$\pm$\bf{0.2}&10.096$\pm$0.7& \bf{5.057}$\pm$\bf{1.2}&6.473$\pm$1.5\\
        time total [ms]& 357.8$\pm$2520.3&\bf{18.2}$\pm$\bf{9.8}&96.8$\pm$142.1 & 3084.6$\pm$2938.0&\bf{152.5}$\pm$\bf{89.1}&4441.6$\pm$10194 & 335.1$\pm$167.2&\bf{19.3}$\pm$\bf{7.8}& 4147.8$\pm$2674.6&\bf{307.9}$\pm$\bf{226.3}\\
        time TO [ms]& 357.8$\pm$2520.3&\bf{13.9}$\pm$\bf{5.7}&91.3$\pm$138.6 & 3084.6$\pm$2938.0&\bf{17.4}$\pm$\bf{14.4}&4236.5$\pm$10135 &335.1$\pm$167.2&\bf{15.9}$\pm$\bf{6.1}& 4147.8$\pm$2674.6&\bf{189.1}$\pm$\bf{170.7}\\
        time CS [ms]& - & 4.2$\pm$4.7&5.5$\pm$5.5& - & 135.1$\pm$76.0&205.2$\pm$107.8& - & 3.4$\pm$2.6 & - & 118.8$\pm$61.0\\
        num. CS& - &2.45$\pm$0.9&3.67$\pm$2.3& - & 3.49$\pm$1.2&6.18$\pm$2.8& - & 2.45$\pm$0.9 & - & 3.49$\pm$1.2\\
        TO SR&0.993&\bf{1.000}&\bf{1.000}&0.863&\bf{1.000}&0.993&0.459&\bf{1.000}&0.735&\bf{1.000}\\
        CFR&0.970&\bf{1.000}&\bf{1.000}&0.725&\bf{1.000}& 0.993&0.459&\bf{1.000}&0.732&\bf{1.000}\\
        CFR given TO S&0.977&\bf{1.000}&\bf{1.000}&0.840&\bf{1.000}&\bf{1.000}&\bf{1.000}&\bf{1.000}&0.995&\bf{1.000}\\

        \hline
\end{tabular}
}
\begin{tabular}{|l|cccccc|}
abbreviation&S &SR &  CFR & TO & LS&CS\\
    \hline
    expanded & success & success rate & collision-free rate & trajectory optimization & line segments & convex sets\\
    \hline
\end{tabular}
\caption{Simulation results of solving the minimum distance and minimum time problems on both the Forest and Franka environments. The table indicates the averages and empirical standard deviations of the statistics across the different road map sizes, and random seeds for the environment and road map construction provided the DRM found a solution.
}
\label{tab:benchmark}
\end{table*}

The simulation experiments are run on an Intel i9-10850K CPU and a NVIDIA GeForce RTX 3090 using the 560.28.03 graphics driver and CUDA 12.6.

We solve each motion planning problem for 10 random seeds of the environment, 10 random seeds for the DRM construction, and 4 different road map sizes, resulting in 400 planning problems for each trajectory optimization approach. The DRM sizes for the Forest environment are 200, 400, 800, and 1600, and for the Franka environment, the sizes are 3000, 6000, 9000, and 12000 nodes. We tuned NTO individually for both environments and problem types. Please refer to \Cref{app:supmatEE} for more specifics on the used parameters. The benchmark results are summarized in \Cref{tab:benchmark}. 

In our experiments, the SCS approaches provide a substantial speedup and boost in the reliability over NTO. We find that, provided DRM success, the SCS approaches never failed, and computed solutions 18.5 times faster in the forest and 15.7 times faster in the franka environment, on average. Given DRM success, NTO produced a collision-free trajectory in 71.5\% of the instances in the Forest environment and 72.8\% of the instances in the Franka environment. We found tuning NTO for the minimum time problem instances in the Forest environment particularly challenging, only achieving a success rate of 45.9\%.

Provided NTO succeeded, it produced equivalent cost trajectories on the Forest minimum distance. On the minimum time instances, it improved on the cost by 7.7\%, provided the optimizer found a solution. For the Franka minimum distance and minimum time instances, it reduced the cost to the SCS approaches by 33.7\%, and 
28\%, respectively.

Inflating two paths and solving the minimum distance problem using LGCS only yielded marginal improvements in the cost of the trajectory, while dramatically increasing the computation time. 

For the Forest environment, the recovery mechanism described in \S\ref{ssec:recovery}, was triggered in 5.6\% of the minimum distance and 0.46\% minimum time problems. In the Franka environment, it was triggered in $40.22\%$ and 21.36\% of the minimum distance and minimum time instances, respectively.

In summary, our pipeline affords a slight increase in trajectory cost, at the benefit of substantially increasing reliability and reducing computation times.

\subsection{Validation on Hardware}\label{ssec:hardware}
\begin{figure*}[t]
    \centering
    \includegraphics[width=0.9\textwidth]{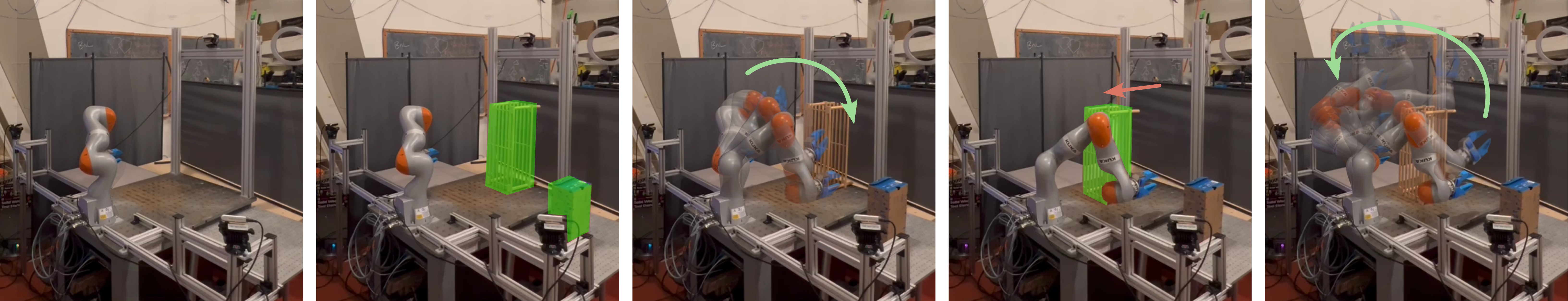}
    \caption{In our hardware experiments, we alternate between planning to the right and the left side of the table. Between plans we place, move and remove obstacles. Starting from the left, first we place the shoe rack and box, highlighted in green, then we run our planner and move the robot to the right side of the table. Next, we slide the shoe rack closer to the robot and run our planner again. The shoe rack forces the arm to move in a higher arc to reach the left side of the table.}
    \label{fig:hardware_experiment}
\end{figure*}
In this section, we validate our pipeline on our system that consists of a KUKA LBR iiwa 7 R800 and three Intel RealSense D415 depth cameras for perceiving changes to the table-top environment shown in \Cref{fig:title_figure} and \Cref{fig:hardware_experiment}. For this setup, we only inflate the shortest path in the DRM and use SCSTrajopt  \cite{marcucci2025biconvex} with Clarabel \cite{goulart2024clarabel} to find good solutions to \eqref{eqn:mintime_prob}. We constrain the velocity between -1 and 1 [$\tfrac{rad}{s}$] and the acceleration between -1 and 1 [$\tfrac{rad}{s^2}$] for all joints.

In our hardware experiments, we alternate between pose targets on the right and the left of the table. Between plans, we modify the environment, as shown in \Cref{fig:hardware_experiment}, by placing or moving obstacles in the center of the table for the robot manipulator to plan around. We approximate the robot collision geometries with 11 boxes, resulting in 105 self-collision pairs. For our hardware experiments, we use an Intel i9-7900X CPU, a Nvidia GeForce 2080Ti GPU using the 555.42.06 driver and CUDA 12.5.  

We implement a simple perception pipeline that uses the Python interface to the depth cameras and down-samples the point clouds to voxels with a side length of 8 cm. We associate a sphere collision geometry that circumscribes each voxel as shown in the center frame of \Cref{fig:title_figure}.

The timing and auxiliary statistics of 15 trajectories with changing obstacle setups are summarized in \Cref{tab:hw_timing}. For additional details please refer to \Cref{app:supmatEE} and the video accompanying this paper that demonstrates the setup.

We find that our entire pipeline ran in 0.82 seconds on average, of which only around 0.12 seconds were spent on constructing and editing the safe sets, and 0.08 seconds were spent running SCSTrajOpt. The planner did not fail during our experiments in this setup. 

\begin{table}[]
    \centering
    \begin{tabular}{lc}
        \toprule 
        Component & Time [ms]\\ \midrule
         perception & 369.7 $\pm$ 16.7\\
IK & 130.1 $\pm$ 79.5\\
DRM & 116.8 $\pm$ 34.8\\
convex sets & 122.5 $\pm$ 43.1\\
SCSTO & 82.8 $\pm$ 67.9\\
overhead & 1.5 $\pm$ 0.6\\
total & 823.4 $\pm$ 122.6\\
         \bottomrule
    \end{tabular}
    \begin{tabular}{lc}
        \toprule 
        Auxiliary Stats. & Value\\ \midrule
         num. line seg. & 2.9 $\pm$ 0.6\\
         num. safe sets &1.93 $\pm$ 0.57\\
         col. checks / set & 121018 $\pm$ 24262\\
         hyperplanes / set &60.7 $\pm$ 14.4\\
         voxels & 359.6 $\pm$ 76.4\\
         col. pairs &4431.2 $\pm$ 916.7\\
         self-col. pairs &105\\
         \bottomrule
    \end{tabular}
    \caption{Statistics (the mean and standard deviation) for the hardware experiments. For the 15 collected trajectories we aggregate the timing information on the left, and auxiliary information on the right. Between each of the trajectories, the obstacle setup was modified similarly to  \Cref{fig:hardware_experiment}.}
    \vspace{-0.5cm}
    \label{tab:hw_timing}
\end{table}







\section{Limitations}\label{sec:limitations}
We discuss four current limitations of our approach.
\subsection{Local and Global Improvements with Nonlinear Solvers}
\ssmps are restricted to searching for trajectories inside of the union of the convex sets. If our method seeds the set construction with a bad path, or our approach for finding solutions to \eqref{opt:closestcollision} results in small sets, it is possible that no high-quality trajectories are covered; the result will potentially have a high cost. For nonlinear trajectory optimization, on the other hand, one can get lucky, and, starting from a bad initial guess, the solver can escape to a better local minimum and still find a good trajectory. We believe that this is reflected in the lower trajectory costs in \Cref{tab:benchmark} when the solver succeeded. We hoped to recover this behavior for the minimum distance problem instances by inflating multiple paths, which indeed created more intricately connected graphs of convex sets between the individual SCS. Unfortunately, resulting improvements on path length were only marginal, even when solving the trajectory optimization to global optimality with the LGCS approach. 

The low trajectory costs of NTO, although, demand scrutiny. Depending on the use case, a system failure could have a high cost (e.g. human intervention) which could make the expected cost of NTO substantially worse than that of the SCS approaches. 

\subsection{GPU Memory}
Currently, the resolution of the voxel maps we can handle is quite limited by the GPU memory. The layout of our collision checker requires checking the robot for collisions against each voxel for all configurations in parallel. Hence, increasing the resolution of the voxel map increases the memory requirements for the collision checker cubically. We aim to mitigate this in the future by representing the environment using a signed distance field \cite{cao2010parallel, chen2022gpu, millane2024nvblox}.   

\subsection{Scaling to Higher Dimensions}
Our experiments only investigate systems up to 7 degrees of freedom. We have yet to investigate how our method works in higher dimensions, e.g. in mobile or bimanual systems. In particular, it remains to be seen up to what dimension our approaches for constructing dynamic roadmaps and configuration-space sets remain effective. 

\subsection{Inverse Kinematics}
Our hardware experiments only featured simple inverse kinematics problems. The obstacles were placed such that they would not overlap with the target end effector pose. We found our inverse kinematics approach to be a weak point of the pipeline, and have not tested its limits in this paper. In the future, we aim to improve this with an approach that is more tailored to the individual robot we are running our pipeline on, e.g. by using analytic inverse kinematics \cite{faria2018position}.

\pwcomment{Should we add a comment about how we are using a very general formulation for NTO and general solvers. Perhaps using something like ALTRO, or similar, wouldve changed the results?}
\section{Conclusion}\label{sec:conclusion}

We have proposed a massively parallelizeable algorithm, EI-ZO, for inflating collision-free line segments to probabilistically collision-free polytopes in robot configuration space. We have demonstrated this algorithm in a pipeline that enables motion planning in changing environments with perception in the loop. The motion planning pipeline leverages DRMs to rapidly find collision-free piecewise linear paths, inflates the paths with EI-ZO, and leverages \ssmps to optimize the final trajectory while eliminating collisions from the sets with candidate trajectories where necessary.

Our pipeline generates the configuration-space convex sets fast enough such that it enables using \ssmps in changing configuration spaces for the first time. As a result, the user can encode nontrivial costs, such as traversal time, and constraints, such as velocity and acceleration bounds, and still profit from the speed and reliability of these approaches. 

Our experimental evaluation demonstrates that generating sequences of convex sets and using corresponding trajectory optimization approaches is not only fast but also reliable. On our simulation benchmark, this approach did not fail provided the DRM found a solution, which resulted in a 27.9\% increase in reliability and a 17.1 times speedup over the nonlinear trajectory optimization baseline. Currently, this comes with the trade-off that this baseline tends to produce trajectories with a lower cost, provided the solver finds a solution.

In conclusion, our procedure is complementary to advances in sampling-based motion planning and advances in \ssmps.
We hope to leverage the former to improve path generation, e.g. via \cite{thomason2024motions,wilson2024nearest, ramsey2024collision}, and the latter to expand the toolbox of supported costs and constraints for the user. In the future, we seek to leverage these advances and deploy our pipeline in practical pick-and-place and higher degree-of-freedom applications such as mobile manipulation.

\pwcomment{Mention not building sets from scratch? mention scaling to higher dimensions? Bimanual?}

\section*{Acknowledgments}
We thank Tobia Marcucci for his insights and for providing the code for SCSTrajopt. We further thank Rebecca H. Jiang, Alexandre Amice, Nicholas Pfaff, Evelyn Fu, Thomas Cohn, Hongkai Dai, Xuchen Han, and the Mobile Manipulation Team at the Toyota Research Institute for many fruitful discussions, help with the software development, and assistance with the hardware experiments. We are grateful for the funding provided by the Toyota Research Institute.

\bibliographystyle{IEEEtran}
\bibliography{biblio}
\appendix
\section{Appendix}\label{appendix}

\subsection{Distance Function Convexity}\label{app:dist_convex}
\textit{Proof}: Let $\calC\subseteq\mathbb{R}^n$ be a convex set, $\lambda\in[0,1]$, and let $x_1,x_2\in\mathbb{R}^n$. Then, we directly confirm that Jensen's inequality holds.
\begin{align}
\begin{split}    
    \lambda{\bf dist}_\calC(x_1) + (1-\lambda){\bf dist}_\calC(x_2) \\
    = \lambda||x_1-z_1|| + (1-\lambda)  ||x_2-z_2||
\end{split}
\end{align}

where $z_1,z_2\in\calC$ are any minimizers of the respective distances to $x_1$ and $x_2$. Observe that 
\begin{align}
\begin{split}
    \lambda||x_1-z_1|| + (1-\lambda)  ||x_2-z_2|| \\\geq ||\lambda x_1 +(1-\lambda)x_2 - (\lambda z_1 +(1-\lambda)z_2)||,
\end{split}
\end{align}
by the triangle inequality, and due to the convexity of $\calC$, we have $(\lambda z_1 +(1-\lambda)z_2)\in\calC$. Hence, it holds
\begin{align}
\begin{split}
    ||\lambda x_1 +(1-\lambda)x_2 - (\lambda z_1 +(1-\lambda)z_2)||\\\geq {\bf dist}_\calC(\lambda x_1 +(1-\lambda)x_2),
\end{split}
\end{align}
showing that Jensen's inequality is satisfied. \qed




\subsection{DRM Construction}\label{app:drm_construction}

Our offline construction of the DRM closely follows \cite{cheng2023motion}. First, we sample a batch of configurations uniformly in the collision-free configuration space of our system to build up our node map $\calM_n$. Next, we construct our node adjacency map $\calM_a$ from the sampled nodes. We limit the number of neighbors for each node to $k$, and only add edges between two configurations if both their $L_2$ distance in configuration space is smaller than $d_\text{CS}$ and the distance in task space of a selected frame is less than $d_\text{TS}$. Then we ensure that the adjacency matrix is symmetric by adding missing edges.

To build the collision map $\calM_c,$ we discretize the task space of the robot into voxels. We use a sphere with radius $\tfrac{\sqrt{3}}{2}s$, where $s$ is the side length of a voxel, that circumscibes each voxel for collision checking. For each voxel, we then check for collision with \textit{each} node of the roadmap, and store the ids of all the nodes in the roadmap that this voxel is in collision with. The resultant mapping between voxels to nodes in collision then serves as our collision map, and this allows for fast look-up of collisions during online planning. 

\subsection{Polygonal Shortest Paths through Sequences of Convex Sets}\label{app:LSCS}

\begin{subequations}\label{eqn:LSCS}    

Given a sequence of convex sets $\calP_{1:M}$ such that successive sets intersect, $\calP_i\cap\calP_{i+1} \neq\emptyset$, then the LSCS second-order cone program is
\begin{align}
    \minz_{v_{0\leq i\leq M+1}} &\alignspacing \sum_{i = 1}^{M}||v_i - v_{i+1}||_2,\\
    \subjectto& \alignspacing v_1 = s,\\
    &\alignspacing v_{M+1} =g,\\
    &\alignspacing v_i = v_{i+1} \text{ for } i=1, \dots ,M, \\
    &\alignspacing v_i\in \calP_i, \text{ for } i=1, \dots ,M,\\
    &\alignspacing v_i\in \calP_{i-1}, \text{ for } i=2, \dots ,M,
\end{align}
\end{subequations}
where $v_i$ are the knot points of the polygonal path.

\subsection{Supplementary Information on Experimental Evaluation} \label{app:supmatEE}

In this section, we collect additional statistics and parameter values that we used in our experiments. This section is split into two parts. First, we cover the simulation experiments, and then the hardware experiments.

\noindent\textbf{Simulation} The simulation experiments are run on a PC running Ubuntu 22.04 with an Intel i9-10850K CPU and an NVIDIA GeForce RTX 3090 using the 560.28.03 graphics driver and CUDA 12.6.

With this setup, our collision checker has a throughput of around 19.9 million configurations per second in the Forest environment and 8.6 million configurations per second in the Franka environment when using a batch size of two million configurations.

We found the greatest success solving all NTO instances in the Forest environment using SNOPT. We employ $N_c = 20$ and $N_c= 100$ for the minimum distance and minimum time problems, respectively. In the Franka environment, we found the greatest success solving the minimum distance problems with SNOPT and the minimum time instances with IPOPT. In both cases $N_c = 150$ yielded a good trade-off between computation time and the resulting trajectories being collision-free when the solver reported success. The parameters employed in the DRM construction for both the Forest and the Franka environments are summarized in \Cref{tab:DRMparamsfranka}. The employed EI-ZO parameters are summarized in \Cref{tab:EI-ZOparamsforest} for the Forest environment, and in \Cref{tab:EI-ZOparamsfranka} for the Franka environment. With these settings, EI-ZO would typically pass the unadaptive test in less than 10 iterations. For the lazy collision checking, we used a step size of 0.1 [m] or 0.1 [rad] in both environments, respectively.

\begin{table}[]
    \centering
    \begin{tabular}{lcc}
        Parameter & Symbol & Value \\
        \hline 
        admissible uncertainty &$\delta$ & 0.05\\
        admissible fraction in collisions&$\varepsilon$ & 0.01\\
        number of samples to update &$N_p$ & 1000\\
        max number of faces per iteration &$N_f$& 10\\
        number of mixing steps for sampling&$N_\text{ms}$& 30\\
        step back&$\Delta_\text{max}$ & 0.01\\
    \end{tabular}
    \caption{Employed parameters in the Forest environment for EI-ZO.}
    \label{tab:EI-ZOparamsforest}
\end{table}

\begin{table}[]
    \centering
    \begin{tabular}{lcc}
        Parameter & Symbol & Value \\
        \hline 
        admissible uncertainty &$\delta$ & 0.005\\
        admissible fraction in collision&$\varepsilon$ & 0.005\\
        number of samples to update &$N_p$ & 10000\\
        max number of faces per iteration &$N_f$& 10\\
        number of mixing steps for sampling&$N_\text{ms}$& 60\\
        step back&$\Delta_\text{max}$ & 0.01\\
    \end{tabular}
    \caption{Employed parameters in the Franka environment for EI-ZO.}
    \label{tab:EI-ZOparamsfranka}
\end{table}

\begin{table}[]
    \centering
    \begin{tabular}{lcc}
        Parameter & Symbol & Value \\
        \hline 
        max. task space distance &$d_\text{TS}$ & 10 [m]\\
        max. configuration space distance & $d_\text{CS}$&10 [m]\\
        max. neighbors &$k$& 10\\
        offline voxel length &$s$&0.06 [m]
    \end{tabular}
    \caption{Employed parameters in the Franka and Forest environments for constructing the DRMs.}
    \label{tab:DRMparamsfranka}
\end{table}



    


\noindent\textbf{Hardware} For our hardware experiments, we use an Intel i9-7900X CPU, a Nvidia GeForce 2080Ti GPU using the 555.42.06 driver and CUDA 12.5. With this setup, our collision checker has a throughput of around 6.9 million configurations per second when using a batch size of 2 million. To generate the PWL paths, we employed a DRM with 35000 nodes. We used the same EI-ZO parameters as in \Cref{tab:EI-ZOparamsfranka}, except for $N_f$, which was increased to 20. In those experiments, the recovery mechanism from \S\ref{ssec:recovery} was never triggered. See \Cref{tab:IIWADRMparams} for the employed parameters during the DRM construction. For the lazy collision checking, we used a step size of 0.1 [rad].

\begin{table}[]
    \centering
    \begin{tabular}{lcc}
        Parameter & Symbol & Value \\
        \hline 
        max. task space distance &$d_\text{TS}$ & 0.45 [m]\\
        max. configuration space distance & $d_\text{CS}$&4.5 [rad]\\
        max. neighbors &$k$& 10\\
        offline voxel length &$s$&0.06 [m]
    \end{tabular}
    \caption{Employed DRM construction parameters for the hardware expermients.}
    \label{tab:IIWADRMparams}
\end{table}

\end{document}